\documentclass[journal]{IEEEtran}
\makeatletter
\def\ps@headings{%
	\def\@oddhead{\mbox{}\scriptsize\rightmark \hfil \thepage}%
	\def\@evenhead{\scriptsize\thepage \hfil \leftmark\mbox{}}%
	\def\@oddfoot{}%
	\def\@evenfoot{}}
\makeatother
\usepackage{multirow}
\usepackage{xcolor}
\usepackage{graphicx,epstopdf}
\usepackage{amsmath,amsthm,amssymb,amsfonts}
\usepackage{tabularx,pdfcomment}
\usepackage{graphicx,color,multirow,amsmath}
\usepackage{cite}
\usepackage{makecell}

\usepackage{tikz}
\usetikzlibrary{arrows.meta,positioning,calc,fit}

\usepackage{subfigure}
\usepackage{hyperref}
\usepackage{url}
\usepackage{cases}
\usepackage{bbm}
\usepackage{bm}
\usepackage{booktabs}
\usepackage{algpseudocode}
\usepackage{algorithmicx}
\usepackage{algorithm}

\usepackage[square, comma, sort&compress, numbers]{natbib}

\theoremstyle{plain} 
\theoremstyle{plain} 
\theoremstyle{plain} 
\theoremstyle{plain} 
\theoremstyle{plain} 
\theoremstyle{plain}

\renewcommand\appendixname{Appendix}

\begin{document}
\title{Large Vision Model–Guided Masked Low-Rank Approximation for Ground-Roll Attenuation}

	
	\author{Jiacheng~Liao,~\IEEEmembership{Graduate Student Member,~IEEE,}
		Feng~Qian,~\IEEEmembership{Senior Member,~IEEE,} Yongjian~Guo, Ziyin~Fan
		                   
		\thanks{This work was supported by the National Natural Science Foundation of China under Grant No. 4254168, and Grant No. 42130812. (\textit{Corresponding author: Feng~Qian.}) }
		\thanks{Jiacheng~Liao, Feng~Qian, and Yongjian~Guo are with the School of Information and Communication Engineering, Center for Information Geoscience, University of Electronic Science and Technology of China, Chengdu, 611731 China (e-mail: jiachengliao004@gmail.com;~fengqian@uestc.edu.cn;~gesila0601@gmail.com).}
		\thanks{Ziyin~Fan is with the School of Resources and Environment, Center for Information Geoscience, University of Electronic Science and Technology of China, Chengdu, 611731 China (e-mail:~yingf946@gmail.com).}
		
		\label{key}
		}

\markboth{IEEE TRANSACTIONS ON GEOSCIENCE AND REMOTE SENSING}%
{LIAO \MakeLowercase{\textit{et al.}}: LVM-LRA for Ground-Roll Attenuation}

\maketitle

\begin{abstract}
Ground roll is a common type of coherent noise in seismic records, and its attenuation remains challenging due to its substantial overlap with useful reflections in localized regions. Existing attenuation methods can be broadly classified into global and local categories according to whether ground-roll-contaminated regions are explicitly identified. Global methods, however, typically impose uniform attenuation on both contaminated and uncontaminated regions, which may result in signal leakage or distortion of reflections. By contrast, local methods restrict attenuation to contaminated regions and are therefore less prone to unnecessary modification of clean areas. However, their performance is often limited by manually designed or simplistic model-based mask estimation strategies. To address these limitations, we propose a large vision model-guided masked low-rank approximation (LVM-LRA) framework for ground-roll attenuation. Within this framework, a promptable LVM is first employed to identify ground-roll-dominant regions in seismic records through multimodal prompting and to generate accurate, fine-grained masks. The estimated masks are then incorporated into an LRA model for ground-roll attenuation. A global low-rank constraint is imposed on the reflection component to preserve event continuity, whereas a mask-guided local low-rank constraint is imposed on the ground-roll component so that its separation is confined to the masked regions. An iterative optimization algorithm based on the alternating direction method of multipliers (ADMM) is further developed to solve the resulting model efficiently. Experiments on synthetic and field datasets demonstrate that the proposed method achieves more effective ground-roll attenuation and better suppresses signal leakage than the baseline methods.
\end{abstract}


\IEEEpeerreviewmaketitle

\section{Introduction}
\label{sec:intro}
Ground roll is a dominant form of coherent noise in land seismic acquisition and is typically characterized by low frequency, low apparent velocity, high amplitude, and strong dispersion \citep{yang2023deep, li2025inr}. Its attenuation remains challenging for two principal reasons. First, owing to its dominant energy, ground roll can contaminate seismic records across offsets, mask weak reflection events, and consequently degrade subsequent imaging and interpretation. Second, its energy often overlaps substantially with useful reflections in the frequency domain \citep{henley2003coherent, leite2009seismic, li2018deep}. Ground-roll attenuation should therefore be viewed not merely as a filtering task, but more fundamentally as a mixed-wavefield separation problem, in which effective noise suppression must be achieved while preserving weak reflection structures to the greatest extent possible \cite{shen2009seismic}.

To address the separation problem described above, current ground-roll attenuation methods can be broadly divided into two categories depending on whether attenuation is applied globally or locally: global suppression methods and local separation methods \cite{cary2009eigenimage}. Global suppression methods treat the contaminated gather as a whole, without explicitly identifying ground-roll-dominant regions beforehand, and can be broadly divided into three groups.  The first group consists of transform-domain methods, which assume that ground roll can be separated from reflections in predefined global domains according to frequency, apparent velocity, dip, or multiscale features. Representative techniques include f-k filtering \cite{wiggins1966wk}, S-transform-based methods \cite{askari2008sxfk}, wavelet transforms \cite{welford2004wavelet}, curvelet transforms \cite{liu2018ssct}, and radial trace transform \cite{henley2003coherent}. The second group comprises global structured modeling methods, which exploit the global predictability, coherence, or low-rank characteristics of ground roll and reflections for separation. Typical examples include matching filtering \cite{jiao2015nmf}, singular value decomposition (SVD)-based filtering \cite{cary2009ground}, and eigenimage-based global decomposition \cite{liu1999ground}. However, these methods apply a uniform global suppression strategy to the entire shot gather, which inevitably damages useful signals in reflection-dominated regions.


The third category includes deep learning (DL)-based global mapping methods, which learn a data-driven mapping over the entire shot gather from contaminated inputs to desired outputs, such as denoised records or separated components. Depending on the availability of reference targets during training, these methods can be broadly divided into supervised and unsupervised paradigms. Supervised methods, including CNN-based approaches \citep{liu2020cnn3d,li2022ffcnn}, GAN-based approaches \cite{yuan2020gan}, and CycleGAN-based approaches \cite{li2021cyclegan}, typically rely on paired or approximately paired training samples. When sufficiently representative training data are available, they can effectively model complex nonlinear relationships between ground roll and reflections, often achieving strong attenuation performance. Unsupervised methods have been introduced to reduce the reliance on labeled data, but effective ground-roll attenuation often requires going beyond purely data-driven networks by incorporating physical mechanisms into the learning process, such as frequency constraints and moveout-assisted separability enhancement \cite{liu2023unsupgr}, self-supervised objectives or similarity-informed constraints \cite{liu2022sisl,liu2022ns2ns,xu2023dnr}, and implicit neural representations for label-free signal modeling \cite{li2025inr}. However, DL methods share the same limitation as the first two categories of global methods, in that they also inevitably cause damage to useful signals outside noise-dominant regions.

Local separation methods exploit the spatially nonuniform and locally dominant nature of ground-roll contamination by focusing on local regions rather than processing the entire shot gather as a whole. Representative methods include local time-frequency transform (LTFT) \cite{liu2010ltft}, local band-limited orthogonalization (LO) \cite{chen2015lo}, adaptive ground-roll attenuation (AGRA/AGORA) \cite{ali2019agora}, local nonlinear filtering \cite{yuan2022lnf}, localized eigenimage \cite{cary2009eigenimage}, SVD filtering \cite{porsani2010svd}, and local low-rank decomposition methods \cite{kocon2011lwd}. However, the performance of local separation methods strongly depends on the accuracy of masks estimated by manually defined or simple model-based strategies. In practice, such strategies often fail to provide sufficiently accurate masks, as verified in \cite{sun2019ground}.

Motivated by the strong segmentation capability of large vision model (LVM) \citep{liu2023clip, wang2024mcpl, khattak2023maple}, the core idea of this study is to replace the simple mask estimation strategies used in local separation methods with an LVM, thereby enabling accurate identification of ground-roll-dominant regions and mitigating the primary source of performance degradation \cite{sun2019ground}. Furthermore, ground roll has been shown to exhibit an approximately low-rank structure in the frequency domain within the masked regions \cite{cary2009ground}. On this basis, a local low-rank approximation (LRA) model is developed to constrain the ground-roll component within the masks. Meanwhile, to preserve weak reflection events, a global low-rank constraint in the frequency domain is imposed on the reflection component \cite{oropeza2011simultaneous}. 

As a result, wavefield separation between ground roll and useful signals is achieved through iterative optimization. Benefiting from the capability of the LVM, the proposed LVM-LRA method exhibits several desirable properties. 1) Minimal signal leakage: The LVM enables accurate mask estimation, which helps reduce the leakage of useful signals in non-noise regions. 2) Label-free learning: Both the LVM and LRA operate without manual annotations and can be trained and adapted in a label-free manner. 3) Scalability and generalization: With prompt-based design, the LVM can be flexibly adapted to different survey areas, offering good scalability and generalization capability. 4) Physical interpretability: Since the LRA is derived from a physical model, its formulation and parameters are physically meaningful and therefore interpretable.

\begin{itemize}
\item[1)] \textit{LVM-guided explicit ground-roll localization:}
CLIPSeg is adopted as the LVM backbone, as it provides a prompt-driven, zero-shot segmentation mechanism for explicit identification of ground-roll-dominant regions in seismic gathers. Specifically, CLIPSeg exploits semantic and visual priors through multimodal prompting \cite{cao2025personalizing}, including morphological prompts, physical-property prompts, and a visual exemplar. To the best of our knowledge, this study represents the first attempt to introduce an LVM for mask identification in ground-roll attenuation.

\item[2)] \textit{Mask-guided low-rank decomposition for region-aware ground-roll separation:}
We propose a mask-guided low-rank decomposition model formulated for wavefield separation with dual low-rank regularization. In this model, a mask-constrained local low-rank regularization term in the frequency domain is imposed on the ground-roll component, whereas a global low-rank regularization term in the frequency domain is imposed on the useful signal component.

\item[3)] \textit{Efficient optimization and effective attenuation performance:}
An ADMM-based optimization strategy is developed for the proposed mask-guided low-rank model, where the coupled objective is decomposed into tractable subproblems with efficient updates \cite{li2020superpixel}. Experiments on synthetic and field datasets show that the proposed framework achieves effective ground-roll attenuation and better suppresses signal leakage than baseline methods.
\end{itemize}

The remainder of this article is organized as follows.
Section~\ref{sec:method} presents the problem statement and semantic-guided inverse formulation.
Section~\ref{sec:stage1} introduces the semantic prior extraction stage using a promptable vision model.
Section~\ref{sec:stage2} describes the mask-conditioned low-rank reconstruction and ADMM solver.
Section~\ref{sec:experiment} details the datasets, baselines, and evaluation protocol.
Section~\ref{sec:results} presents quantitative and qualitative results, including leakage analysis and spectral validation.
Finally, Section~\ref{sec:conclusion} concludes the article.

\begin{figure*}[t!]
	
	\vspace{-0.4cm}
	\centering
	\includegraphics[width=1\textwidth,  trim=15 15 15 15,
  clip]{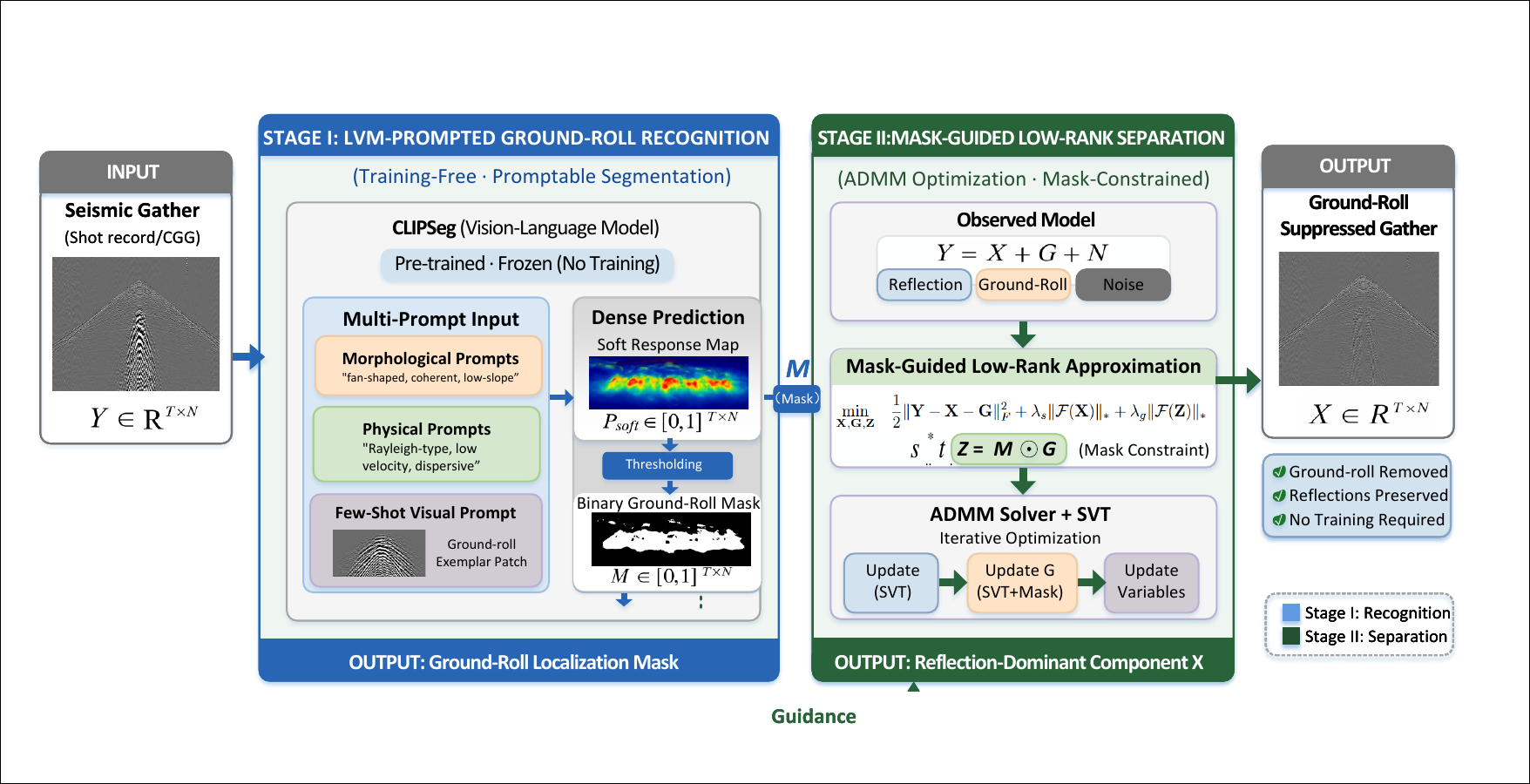}
	\caption{Schematic diagram of the proposed LVM-LRA model.}
	\label{Fig:syn1}
\end{figure*}

\section{Problem Statement and Formulation}
\label{sec:method}

Ground-roll attenuation is formulated as an additive signal decomposition problem. The observed seismic gather $\mathbf{Y} \in \mathbb{R}^{T \times X}$ is modeled as the superposition of the useful image, the ground-roll component, and residual noise:
\begin{equation}
\mathbf{Y} = \mathbf{X} + \mathbf{G} + \mathbf{N},
\end{equation}
where $\mathbf{X} \in \mathbb{R}^{T \times X}$ denotes the useful image, $\mathbf{G} \in \mathbb{R}^{T \times X}$ denotes the ground-roll component, and $\mathbf{N}$ represents the residual noise. Recovering $\mathbf{X}$ and $\mathbf{G}$ from $\mathbf{Y}$ is inherently ill-posed, since the decomposition is nonunique in the absence of additional prior information \cite{WEN2025}. A common strategy is to introduce regularization and formulate the problem as
\begin{equation}
\label{equ:total}
\min_{\mathbf{X}, \mathbf{G}}
\frac{1}{2}\|\mathbf{Y} - \mathbf{X} - \mathbf{G}\|_F^2
+ \lambda_{\mathrm{s}} R(\mathbf{X}) + \lambda_{\mathrm{g}} R(\mathbf{G}),
\end{equation}
where $R(\mathbf{X})$ and $R(\mathbf{G})$ encode prior assumptions on the useful image and ground-roll components, respectively, and $\lambda_s>0$ and $\lambda_g>0$ are the corresponding regularization parameters. However, as pointed out in \cite{porsani2010svd}, ground roll is typically dominant only in localized regions of the record. Imposing a global constraint on $\mathbf{G}$ is therefore not well justified from a modeling perspective and often leads to excessive leakage of useful image content in regions not dominated by ground roll.

Motivated by \citep{9738631,sun2019ground,oliveira2020self,welford2004wavelet}, we introduce a binary spatial mask $\mathbf{M}\in\{0,1\}^{T\times X}$ to explicitly characterize the ground-roll-dominant region. A mask-supported ground-roll component is then defined as $\mathbf{Z}=\mathbf{M}\circ\mathbf{G}$, where $\circ$ denotes the Hadamard product. Accordingly, based on the global low-rank structure of the useful image and the mask-supported local low-rank structure of ground roll in the frequency domain, corresponding rank-based regularization terms are imposed on $\mathbf{X}$ and $\mathbf{Z}$, respectively. Specifically, the nuclear norm is widely used as a surrogate for the rank function $\mathrm{rank}(\cdot)$ to promote low-rank solutions \cite{huang2018joint}. Accordingly, \eqref{equ:total} can be reformulated as
\begin{equation}
\begin{aligned}
\min_{\mathbf{X},\mathbf{G},\mathbf{Z}} \quad
& \frac{1}{2}\|\mathbf{Y}-\mathbf{X}-\mathbf{G}\|_F^2
+ \lambda_s\|{\mathcal{F}(\mathbf{X})}\|_*
+ \lambda_g\|{\mathcal{F}(\mathbf{Z})}\|_* \\
\text{s.t.}\quad
& \mathbf{Z}=\mathbf{M}\circ\mathbf{G}.
\end{aligned}
\label{equ:stage_model}
\end{equation}
where $\mathcal{F}(\cdot)$ denotes the fast Fourier transform (FFT) operator and $\|\cdot\|_*$ denotes the nuclear norm. This formulation leads naturally to a two-stage framework. In Stage I, the ground-roll-dominant region is identified and the corresponding mask is generated. As described in Section \ref{sec:intro}, this mask is obtained using the proposed LVM-guided strategy. In Stage II, an LRA model is established to separate the useful image from the ground-roll component. The two stages are described in detail below.

\section{Stage I: CLIPSeg-Based Multimodal Prompting for Ground-Roll Masking}

Stage~I aims to generate a ground-roll mask for the subsequent mask-guided separation. This section first introduces the CLIPSeg-based mask generator in Section \ref{Gen} and then presents the multimodal prompt design from three perspectives: morphological constraints in Section \ref{sec:prompt_morph}, physical prior knowledge in Section \ref{sec:prompt_phys}, and exemplar guidance in Section \ref{sec:prompt_fewshot}.

\label{sec:stage1}

\subsection{Ground-Roll Mask Generator}
\label{Gen}
Inspired by \cite{liu2023clip}, this study adopts a prompt-driven vision--language segmentation mechanism based on CLIPSeg, which jointly encodes seismic image features and prompt information to generate a dense response map associated with ground-roll patterns. Let $E_Y(\cdot)$ denote the image encoder, $E_P(\cdot)$ the prompt encoder, and $D(\cdot,\cdot)$ the mask decoder \cite{cao2025personalizing}. The prompt-conditioned mask response is expressed as
\begin{equation}
\widetilde{\mathbf{M}} = D\!\left(E_Y(Y),\,E_P(P)\right),
\label{eq:stage1_clipseg}
\end{equation}
where $P$ denotes the multimodal prompt set introduced in Sections~\ref{sec:prompt_morph}--\ref{sec:prompt_fewshot}. The response map $\widetilde{\mathbf{M}}\in[0,1]^{T\times X}$ assigns each time--space sample a confidence value that indicates its association with ground-roll patterns, where larger values correspond to stronger ground-roll dominance \cite{li2024promptad}.

To obtain the binary support used in Stage~II, the response map is binarized as
\begin{equation}
M_{ij} =
\begin{cases}
1, & \widetilde{M}_{ij} > \eta,\\
0, & \text{otherwise},
\end{cases}
\label{eq:stage1_binary}
\end{equation}
where $\eta$ denotes a prescribed threshold. The resulting binary mask $\mathbf{M}$ indicates whether each time--space sample belongs to a ground-roll-dominant region, with $\mathbf{M}_{ij}=1$ indicating ground-roll support and $\mathbf{M}_{ij}=0$ otherwise.

\begin{figure*}[t!]
	
	\vspace{-0.4cm}
	\centering
	\includegraphics[width=1\textwidth, height=0.52\textheight,  trim=15 15 15 15,
  clip]{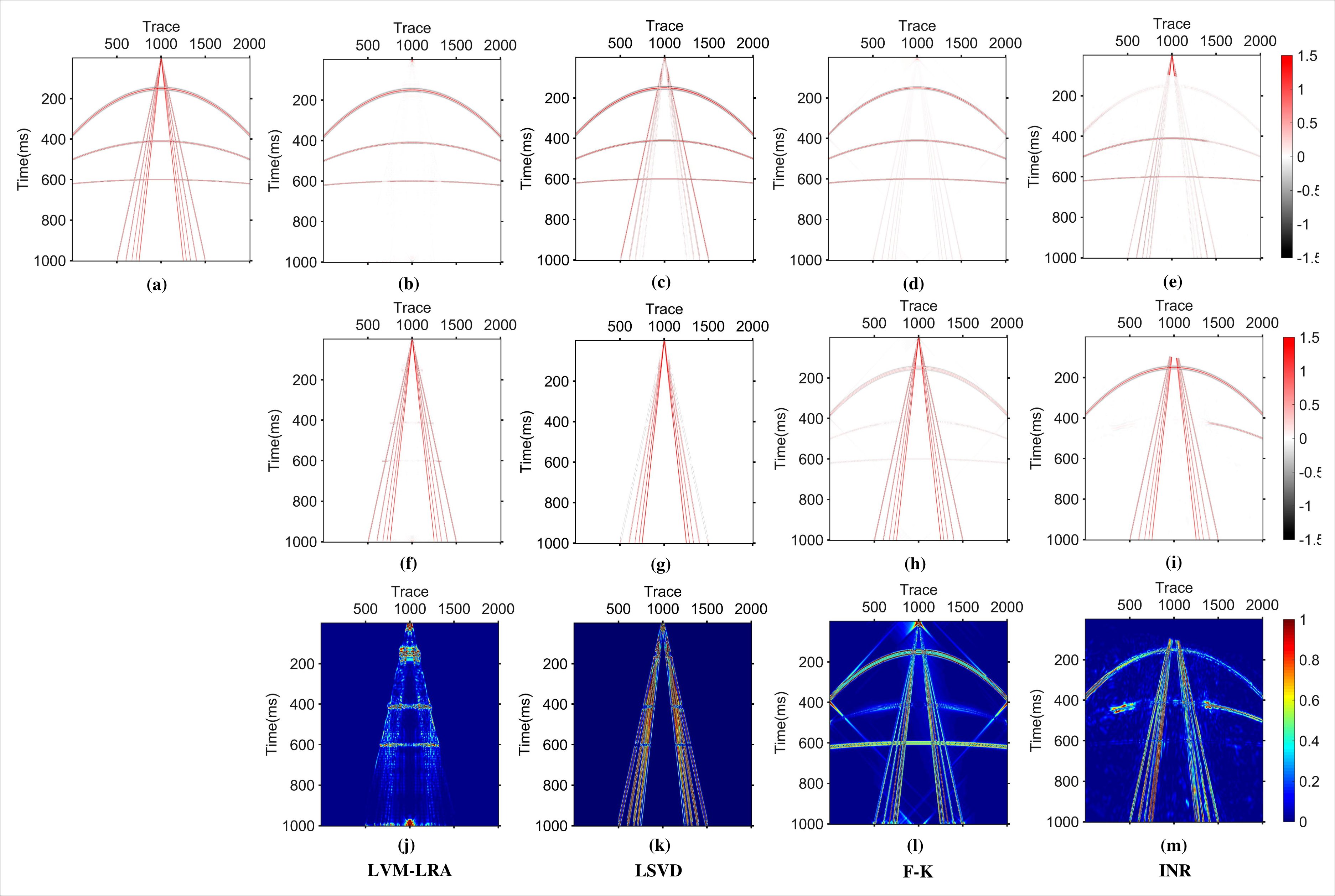}
	\caption{Denoising comparison on synthetic seismic data. (a) Original seismic data. Denoised results, removal noise, and local similarity comparisons using (b)(f)(j) proposed LVM-LRA, (c)(g)(k) LSVD, (d)(h)(l) F-K, and (e)(i)(m) INR.}
	\label{Fig:syn0}
\end{figure*}

\subsection{Prompt I: Morphological Constraints}
\label{sec:prompt_morph}
Ground roll in seismic gathers typically exhibits distinctive visual characteristics, including fan- or cone-shaped energy distributions in near-offset regions, lateral coherence, low apparent slope, and smoothly varying wavefronts \citep{cary2009ground,liu2020should}. Motivated by these phenomenological observations, we encode such visible characteristics as morphological prompts. Here, “morphology” refers to appearance-level features, including geometric shape, spatial continuity, and textural coherence, which guide the vision--language model toward regions visually consistent with ground roll.

Formally, the morphological prompt set is defined as
\begin{equation}
\mathcal{P}_{m}=\left\{p^{(m)}_1,\ldots,p^{(m)}_{N_m}\right\},
\end{equation}
where each prompt captures a salient appearance attribute of ground roll, such as \emph{fan- or cone-shaped energy bands}, \emph{low-slope coherent events}, or \emph{smooth and continuous wavefront textures}. These prompts provide data-agnostic visual guidance, thereby improving the appearance consistency of the generated mask \cite{liu2023clip}.

\begin{figure*}[t!]
	
	\vspace{-0.4cm}
	\centering
	\includegraphics[width=0.75\textwidth,  trim=15 15 15 15,
  clip]{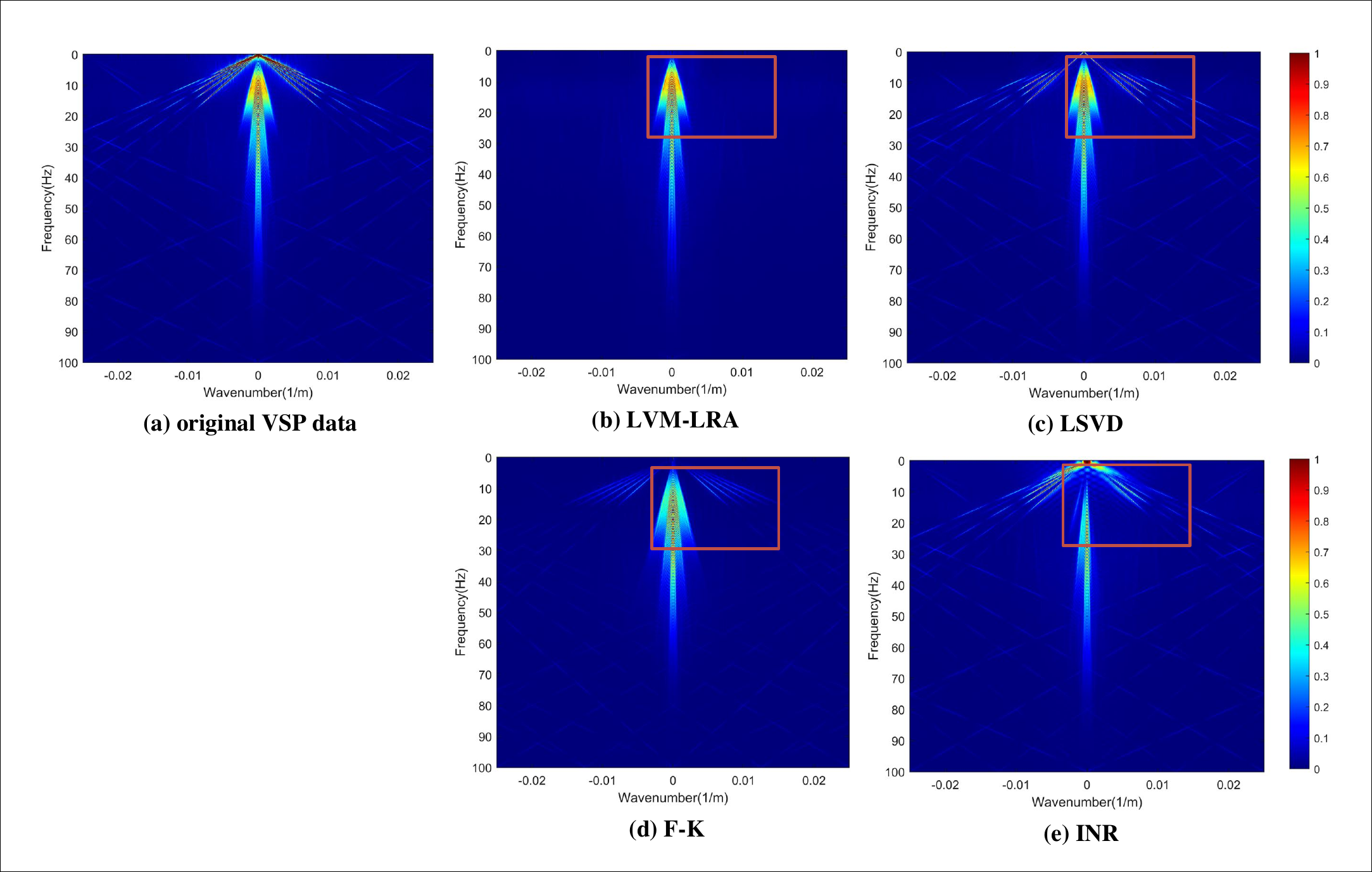}
	\caption{Spectral comparison of the denoised results in Fig.~\ref{Fig:syn0}: (a), (b), (c), (d), and (e), Frequency--wavenumber spectra of Fig.~\ref{Fig:syn0} (a), (b), (c), (d), (e), respectively.}
	\label{Fig:syn1}
\end{figure*}

\subsection{Prompt II: Physical Prior Knowledge}
\label{sec:prompt_phys}

Although morphological prompts capture the visual characteristics of ground roll, appearance information alone may not be sufficiently discriminative in complex seismic scenarios. Other coherent events, such as near-surface reflections, subsurface scattered waves, or locally coherent noise bursts, may exhibit similar local textures or geometric patterns \cite{aghayan2016seismic,pham2022physics}. Furthermore, \cite{xia2014estimation} noted that ground roll can generally be regarded as a Rayleigh-type surface wave propagating along or near the ground surface and characterized by low velocity, low frequency, high amplitude, as well as guided-wave and dispersive behavior. The combination of these physical attributes gives rise to a distinctive wave pattern that helps distinguish ground roll from other coherent events.

Accordingly, a physics-prior prompt set is defined as
\begin{equation}
\mathcal{P}_{p}=\left\{p^{(p)}_1,\ldots,p^{(p)}_{N_p}\right\},
\end{equation}
where each prompt encodes an interpretable physical attribute of ground roll, such as \emph{Rayleigh-type surface-wave propagation}, \emph{low-velocity dispersive behavior}, or \emph{near-surface guided-wave energy}. In practice, $\mathcal{P}_{m}$ and $\mathcal{P}_{p}$ are combined to form the multimodal semantic prompt set $P$ in \eqref{eq:stage1_clipseg}, thereby yielding a more reliable and interpretable ground-roll mask $\mathbf{M}$ in an zero-shot manner \cite{khattak2025learning}.

\subsection{Prompt III: Exemplar Guidance}
\label{sec:prompt_fewshot}

While textual prompts provide high-level semantic guidance, certain structural properties of seismic patterns remain difficult to convey through natural language alone. To complement the textual description, a single exemplar image $p^{(v)}$ is incorporated into the multimodal prompt set $P$ as a visual prompt. This exemplar serves as a lightweight cue for reducing ambiguity under seismic domain shift \cite{khattak2023maple}.

Specifically, a representative seismic image containing ground-roll events is selected as the visual exemplar. It implicitly conveys characteristic spatial patterns of ground roll, such as coherent structures and fan-shaped geometry \citep{wu2025visual, luddecke2022image}. A key advantage of this strategy is that it requires only a single exemplar image and does not involve additional training, thereby keeping the overall pipeline lightweight and easy to deploy. By combining semantic text prompts with a visual exemplar, the proposed CLIPSeg-based masking framework becomes better adapted to seismic imagery while preserving the generalization capability of the pretrained vision--language representations, as shown in Table~\ref{tab:prompt_ablation}.

\begin{figure*}[t!]
	
	\vspace{-0.4cm}
	\centering
	\includegraphics[width=1\textwidth,  trim=2 2 2 2,
  clip]{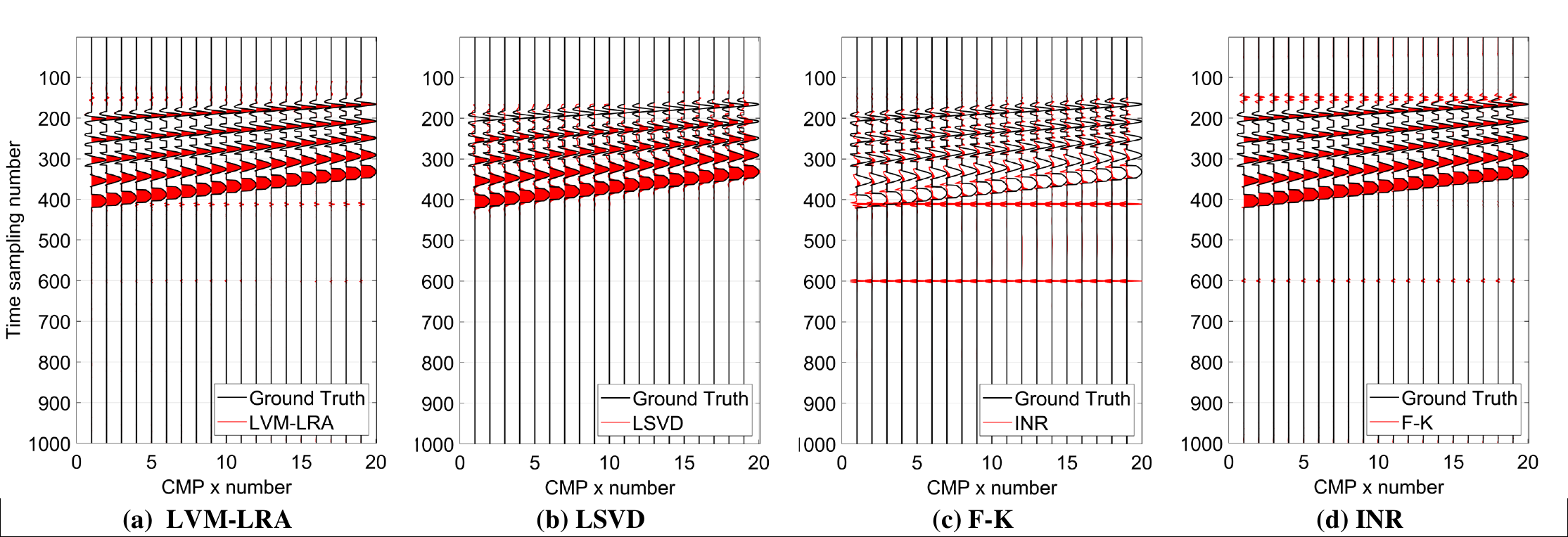}
	\caption{Single-trace comparison of the ground-roll at the 900-918 trace between the original ground-roll and the extracted components by (a) proposed LVM-LRA, (b) LSVD, (c) F-K, and (d) INR.}
	\label{Fig:syn2}
\end{figure*}


\begin{table}[t]
\caption{Ablation study of different prompt combinations for ground-roll mask generation in Stage~I.}
\label{tab:prompt_ablation}
\centering
\renewcommand{\arraystretch}{1.15}
\begin{tabular}{p{0.70\columnwidth} c}
\toprule
\textbf{Prompt design} & \textbf{Mask IoU} \\
\midrule

Generic category-level text prompt (``ground-roll'')
& 0.58 \\

Morphological text prompt only
& 0.71 \\

Physical-property text prompt only
& 0.76 \\

A visual prompt only
& 0.87 \\

Morphological + Physical text prompts
& 0.82 \\

Morphological text + Visual prompt
& 0.89 \\

Physical-property text + Visual prompt
& 0.90 \\

Morphological + Physical + Visual prompts (proposed)
& \textbf{0.92} \\

\bottomrule
\end{tabular}
\end{table}



\section{Stage II: Mask-Guided Low-Rank Approximation for Ground-Roll Suppression}
\label{sec:stage2}

Since ADMM decomposes a complicated optimization problem into a sequence of simpler subproblems, the auxiliary variables $\mathbf{U}$ and $\mathbf{V}$ are introduced to make the objective function separable. By introducing these auxiliary variables, the optimization problem in \eqref{equ:stage_model} can be rewritten as
\begin{equation}
\begin{aligned}
\min_{\mathbf{X},\mathbf{G},\mathbf{Z},\mathbf{U},\mathbf{V}}\quad
& \frac{1}{2}\|\mathbf{Y}-\mathbf{X}-\mathbf{G}\|_F^2
+ \lambda_s\|\mathbf{U}\|_*
+ \lambda_g\|\mathbf{V}\|_* \\
\text{s.t.}\quad
& \mathbf{U}=\mathcal{F}(\mathbf{X}), \quad
\mathbf{Z}=\mathbf{M}\circ\mathbf{G}, \quad
\mathbf{V}=\mathcal{F}(\mathbf{Z}).
\end{aligned}
\end{equation}

The corresponding augmented Lagrangian function is given by
\begin{equation}
\begin{aligned}
\mathcal{L}=&\frac{1}{2}\|\mathbf{Y}-\mathbf{X}-\mathbf{G}\|_F^2
+\lambda_s\|\mathbf{U}\|_*
+\lambda_g\|\mathbf{V}\|_* \\
&+\frac{\rho_1}{2}\|\mathbf{U}-\mathcal{F}(\mathbf{X})+\mathbf{D}_1\|_F^2 \\
&+\frac{\rho_2}{2}\|\mathbf{Z}-\mathbf{M}\circ\mathbf{G}+\mathbf{D}_2\|_F^2 \\
&+\frac{\rho_3}{2}\|\mathbf{V}-\mathcal{F}(\mathbf{Z})+\mathbf{D}_3\|_F^2,
\end{aligned}
\label{eq:augLag}
\end{equation}
where $\rho_1$, $\rho_2$, and $\rho_3$ are penalty parameters, and $\mathbf{D}_1$, $\mathbf{D}_2$, and $\mathbf{D}_3$ are the scaled dual variables associated with the three constraints. The variables are then updated in an alternating manner.

With the other variables fixed, the $\mathbf{X}$-subproblem is written as
\begin{equation}
\mathbf{X}^{k+1}
=
\arg\min_{\mathbf{X}}
\frac{1}{2}\|\mathbf{Y}-\mathbf{X}-\mathbf{G}^{k}\|_F^2
+\frac{\rho_1}{2}\|\mathbf{U}^{k}-\mathcal{F}(\mathbf{X})+\mathbf{D}_1^{k}\|_F^2.
\label{eq:X_sub}
\end{equation}

Under the unitary normalization of the Fourier transform, the update of $\mathbf{X}$ is given by
\begin{equation}
\mathbf{X}^{k+1}
=
\frac{(\mathbf{Y}-\mathbf{G}^{k})+\rho_1\mathcal{F}^{-1}(\mathbf{U}^{k}+\mathbf{D}_1^{k})}{1+\rho_1}.
\label{eq:X_update}
\end{equation}
where $\mathcal{F}^{-1}(\cdot)$ denotes the inverse fast Fourier transform (IFFT).
With $\mathbf{X}^{k+1}$ available, the update of $\mathbf{G}$ is obtained by solving
\begin{equation}
\mathbf{G}^{k+1}
=
\arg\min_{\mathbf{G}}
\frac{1}{2}\|\mathbf{Y}-\mathbf{X}^{k+1}-\mathbf{G}\|_F^2
+\frac{\rho_2}{2}\|\mathbf{Z}^{k}-\mathbf{M}\circ\mathbf{G}+\mathbf{D}_2^{k}\|_F^2.
\label{eq:G_sub}
\end{equation}

The corresponding elementwise solution is given by
\begin{equation}
\mathbf{G}^{k+1}
=
\frac{\mathbf{Y}-\mathbf{X}^{k+1}+\rho_2\,\mathbf{M}\circ(\mathbf{Z}^{k}+\mathbf{D}_2^{k})}
{\mathbf{1}+\rho_2\,\mathbf{M}},
\label{eq:G_update}
\end{equation}
where $\mathbf{1}$ denotes the all-one matrix.

The variable $\mathbf{Z}$ is then updated. The corresponding subproblem is
\begin{equation}
\begin{aligned}
\mathbf{Z}^{k+1}
= &
 \arg\min_{\mathbf{Z}}
\frac{\rho_2}{2}\|\mathbf{Z}-\mathbf{M}\circ\mathbf{G}^{k+1}+\mathbf{D}_2^{k}\|_F^2 \\
&+\frac{\rho_3}{2}\|\mathbf{V}^{k}-\mathcal{F}(\mathbf{Z})+\mathbf{D}_3^{k}\|_F^2.
\label{eq:Z_sub}
\end{aligned}
\end{equation}

The same strategy yields the following closed-form update:
\begin{equation}
\mathbf{Z}^{k+1}
=
\frac{
\rho_2(\mathbf{M}\circ\mathbf{G}^{k+1}-\mathbf{D}_2^{k})
+\rho_3\mathcal{F}^{-1}(\mathbf{V}^{k}+\mathbf{D}_3^{k})
}{\rho_2+\rho_3}.
\label{eq:Z_update}
\end{equation}

The update of $\mathbf{U}$ is obtained by solving
\begin{equation}
\mathbf{U}^{k+1}
=
\arg\min_{\mathbf{U}}
\lambda_1\|\mathbf{U}\|_*
+\frac{\rho_1}{2}\|\mathbf{U}-\mathcal{F}(\mathbf{X}^{k+1})+\mathbf{D}_1^{k}\|_F^2.
\label{eq:U_sub}
\end{equation}

This subproblem corresponds to the proximal mapping of the nuclear norm. Therefore, its solution is given by
\begin{equation}
\mathbf{U}^{k+1}
=
\operatorname{SVT}_{\lambda_1/\rho_1}
\big(\mathcal{F}(\mathbf{X}^{k+1})-\mathbf{D}_1^{k}\big),
\label{eq:U_update}
\end{equation}
where $\operatorname{SVT}_{\lambda_1/\rho_1}(\cdot)$ denotes the singular value thresholding operator with threshold $\lambda_1/\rho_1$.

Similarly, the $\mathbf{V}$-subproblem is given by
\begin{equation}
\mathbf{V}^{k+1}
=
\arg\min_{\mathbf{V}}
\lambda_2\|\mathbf{V}\|_*
+\frac{\rho_3}{2}\|\mathbf{V}-\mathcal{F}(\mathbf{Z}^{k+1})+\mathbf{D}_3^{k}\|_F^2.
\label{eq:V_sub}
\end{equation}

Its solution is obtained in the same manner:
\begin{equation}
\begin{aligned}
\mathbf{V}^{k+1}
=
\operatorname{SVT}_{\lambda_2/\rho_3}
\big(\mathcal{F}(\mathbf{Z}^{k+1})-\mathbf{D}_3^{k}\big).
\label{eq:V_update}
\end{aligned}
\end{equation}

Finally, the scaled dual variables are updated as
\begin{equation}
\begin{aligned}
\mathbf{D}_1^{k+1}
&=
\mathbf{D}_1^{k}+\mathbf{U}^{k+1}-\mathcal{F}(\mathbf{X}^{k+1}), \\
\mathbf{D}_2^{k+1}
&=
\mathbf{D}_2^{k}+\mathbf{Z}^{k+1}-\mathbf{M}\circ\mathbf{G}^{k+1}, \\
\mathbf{D}_3^{k+1}
&=
\mathbf{D}_3^{k}+\mathbf{V}^{k+1}-\mathcal{F}(\mathbf{Z}^{k+1}).
\end{aligned}
\label{eq:dual_update}
\end{equation}

In practice, the algorithm terminates when the maximum number of iterations is reached or when the primal residual satisfies
\begin{equation}
\max\left\{
\begin{array}{l}
\|\mathbf{U}^{k}-\mathcal{F}(\mathbf{X}^{k})\|_F,\\
\|\mathbf{Z}^{k}-\mathbf{M}\circ\mathbf{G}^{k}\|_F,\\
\|\mathbf{V}^{k}-\mathcal{F}(\mathbf{Z}^{k})\|_F
\end{array}
\right\}
\le \varepsilon,
\label{eq:stop}
\end{equation}
where $\varepsilon>0$ is a prescribed tolerance.

\begin{algorithm}[t]
\caption{LVM-LRA for Ground-Roll Attenuation}
\label{alg:lvm_lra}
\footnotesize
\begin{algorithmic}[1]
\Require Observed gather $\mathbf{Y}\in\mathbb{R}^{T\times X}$; threshold $\eta$; ADMM parameters $(\lambda_{\mathrm{g}},\lambda_{\mathrm{s}},\rho_1,\rho_2,\rho_3.,K_{m},\varepsilon)$.
\Ensure Reflection component ${\mathbf{X}}$, ground-roll component ${\mathbf{G}}$

\Statex \textbf{Stage I: LVM-guided mask generation}
\State Construct seismic gather $\mathbf{Y}$
\State Construct multimodal prompt set $P$
\State $\widetilde{\mathbf{M}} \gets D\!\left(E_Y(Y),\,E_P(P)\right)$
\State Binarize $\widetilde{\mathbf{M}}$ using threshold $\eta$ to obtain $\mathbf{M}$

\Statex \textbf{Stage II: Mask-guided frequency-domain low-rank decomposition}
\State Initialize $\mathbf{X}^{0}, \mathbf{G}^{0}, \mathbf{Z}^{0}, \mathbf{U}^{0}, \mathbf{V}^{0}, \mathbf{D}_{1}^{0}, \mathbf{D}_{2}^{0}, \mathbf{D}_{3}^{0} \gets \mathbf{0}$, $k \gets 0$
\While{not converged and $k < K_{m}$}
    \State Update $\mathbf{X}^{k+1}$ using (\ref{eq:X_update})
    \State Update $\mathbf{G}^{k+1}$ using (\ref{eq:G_update}) 
    \State Update $\mathbf{Z}^{k+1}$ using (\ref{eq:Z_update})
    \State Update $\mathbf{U}^{k+1}$ using (\ref{eq:U_update})
    \State Update $\mathbf{V}^{k+1}$ using (\ref{eq:V_update})
    \State Update $\mathbf{D}_{1}^{k+1}$, $\mathbf{D}_{2}^{k+1}$, and $\mathbf{D}_{3}^{k+1}$ using
    (\ref{eq:dual_update})
    \State Check convergence using (\ref{eq:stop})
    \State $k \gets k+1$
\EndWhile

\State ${\mathbf{X}}\gets \mathbf{X}^{k+1}$, \quad ${\mathbf{G}}\gets \mathbf{G}^{k+1}$
\end{algorithmic}
\end{algorithm}

\begin{figure*}[t!]
	
	\vspace{-0.4cm}
	\centering
	\includegraphics[width=1\textwidth, height=0.52\textheight,  trim=15 15 15 15,
  clip]{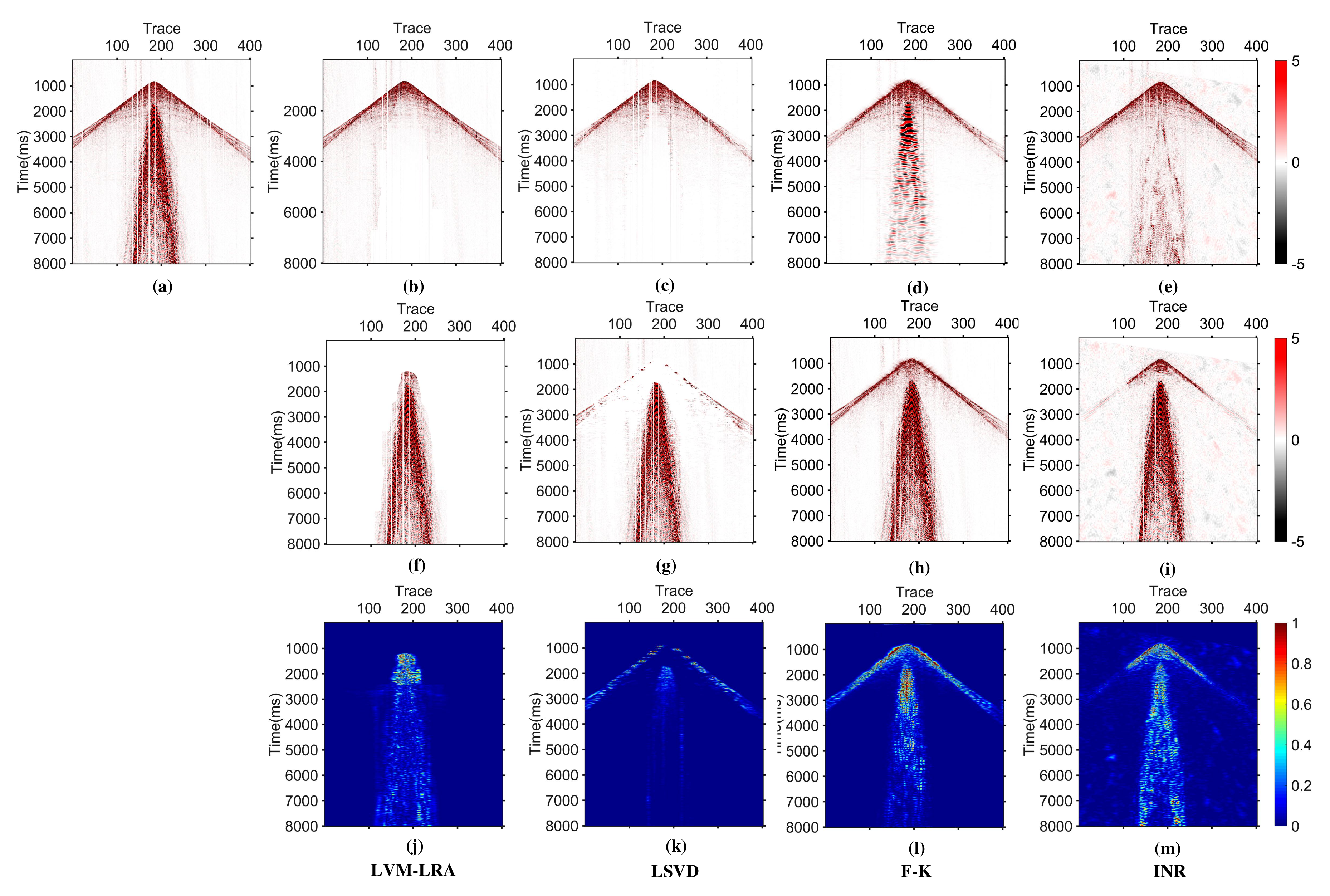}
	\caption{Denoising comparison on the first real seismic data. (a) Original seismic data. Denoised results, removal noise, and local similarity comparisons using (b)(f)(j) proposed LVM-LRA, (c)(g)(k) LSVD, (d)(h)(l) F-K, and (e)(i)(m) INR.}
	\label{Fig:syn3}
\end{figure*}

The dominant computational cost of the Stage II solver arises from the two SVT operations used to update $\mathbf{U}$ and $\mathbf{V}$. Using SVD with target rank $r$, each SVT step requires approximately $\mathcal{O}(TXr)$ operations. The updates of $\mathbf{X}^{k+1}$ and $\mathbf{Z}^{k+1}$ additionally involve FFT/IFFT operations, each with complexity $\mathcal{O}(TX\log(TX))$, whereas the updates of $\mathbf{G}^{k+1}$, the dual variables, and the residuals require only $\mathcal{O}(TX)$. Therefore, the overall computational complexity of the Stage II solver is $\mathcal{O}\bigl(TX\log(TX)+TXr\bigr)$ per iteration. After $K$ iterations, the total computational complexity becomes $\mathcal{O}\bigl(K(TX\log(TX)+TXr)\bigr)$.

With the mask $\mathbf{M}$ fixed from Stage I, the Stage II optimization problem is convex, as it consists of a quadratic data-fidelity term, nuclear-norm regularization terms, and linear equality constraints. Moreover, each subproblem in the proposed ADMM scheme is solved exactly, either in closed form or through the proximal mapping of the nuclear norm. The resulting algorithm therefore falls within the standard ADMM framework for convex splitting problems. According to classical ADMM convergence theory, the generated iterates converge to a primal--dual solution of the underlying convex problem under standard assumptions on the penalty parameters \cite{doi:10.1137/0805023, neal2011distributed}. For brevity, the detailed proof is omitted.

\begin{figure*}[t!]
	
	\vspace{-0.4cm}
	\centering
	\includegraphics[width=0.75\textwidth,  trim=15 15 15 15,
  clip]{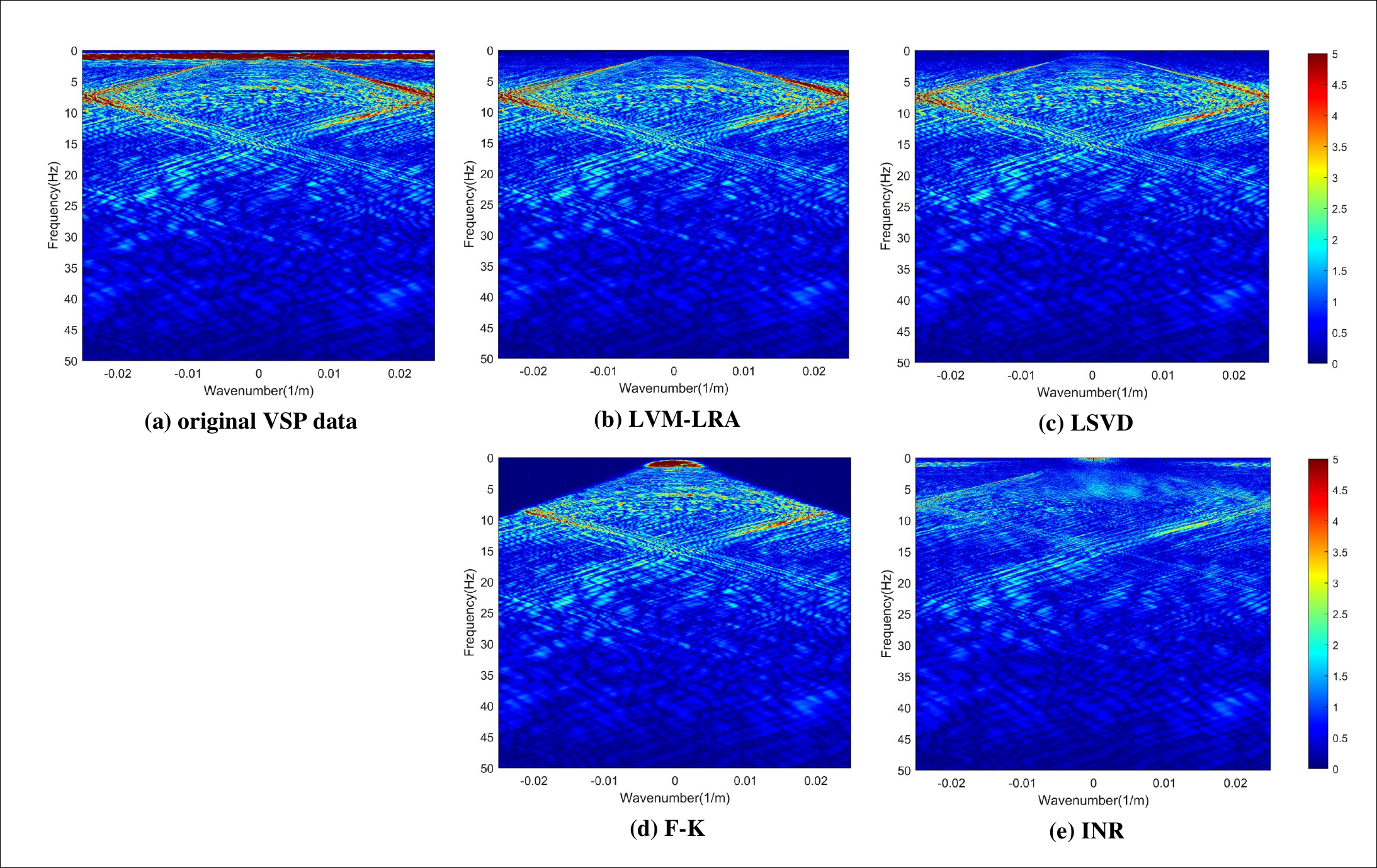}
	\caption{Spectral comparison of the denoised results in Fig. \ref{Fig:syn3}: (a), (b), (c), (d), and (e) Frequency–wavenumber spectra of Fig. \ref{Fig:syn3} (a), (b), (c), (d), and (e), respectively.}
	\label{Fig:syn4}
\end{figure*}

\section{Experimental Setup}
\label{sec:experiment}
\subsection{Baselines}
Based on the literature review in Section~\ref{sec:intro}, three representative seismic noise attenuation methods are selected as baselines. Their core attenuation principles are briefly summarized as follows.

\begin{enumerate}
	\item \textit{Local Singular Value Decomposition (LSVD)\footnote{\url{https://github.com/kianmajl/Image_Denoising_using_SVD}}~\cite{sun2019ground}:}  
    LSVD first estimates a ground-roll mask from the data using local energy and coherence attributes. It then performs singular value decomposition within local sliding windows under the guidance of this mask to reconstruct the dominant coherent ground-roll energy, which is subsequently removed from the input data.
    
	\item \textit{Frequency--Wavenumber (F--K) Filtering\footnote{\url{https://github.com/nicklinyi/seismic_utils}}~\cite{naghizadeh2012seismic}:}    
	F--K filtering exploits differences in apparent velocity among wavefield components. The seismic data are transformed into the frequency--wavenumber domain, where ground-roll energy concentrated in low-frequency and low-velocity regions is attenuated before transformation back to the time--space domain.

	\item \textit{Implicit Neural Representation (INR)\footnote{\url{https://github.com/VITA-Group/INSP}}~\cite{li2025inr}:}   
	INR represents seismic data as a continuous implicit function that maps spatial--temporal coordinates to amplitudes. By fitting this function to the observed data, INR captures the dominant structured components of the seismic signal, whereas components inconsistent with the implicit representation tend to be suppressed during optimization.
\end{enumerate}

\subsection{Evaluation Metrics}
\label{subsec:metrics}

Following the evaluation protocols outlined in~\cite{liang2023reinforcement} and~\cite{Shekhar2025}, signal-to-noise ratio (SNR) and local similarity are adopted as quantitative metrics to assess the denoising performance of the proposed LVM-LRA method against representative state-of-the-art methods. For synthetic datasets, the availability of ground-truth data enables objective evaluation using SNR, which quantifies the relative strength of the preserved seismic signal with respect to residual ground roll and random noise.

For real datasets, where ground truth is unavailable, the performance of the proposed LVM-LRA method is evaluated using local similarity rather than SNR. The formulation and implementation of this metric are described in~\cite{chen2015random} and~\cite{fomel2007local}. Importantly, local similarity is computed at appropriate spatial scales to evaluate signal leakage. This is typically achieved by comparing the denoised result with the extracted ground-roll component. If the extracted noise contains residual useful signal energy, elevated similarity values appear in the corresponding regions, indicating signal leakage and incomplete separation.

\begin{figure*}[t!]
	
	\vspace{-0.4cm}
	\centering
	\includegraphics[width=1\textwidth, height=0.52\textheight,  trim=15 15 15 15,
  clip]{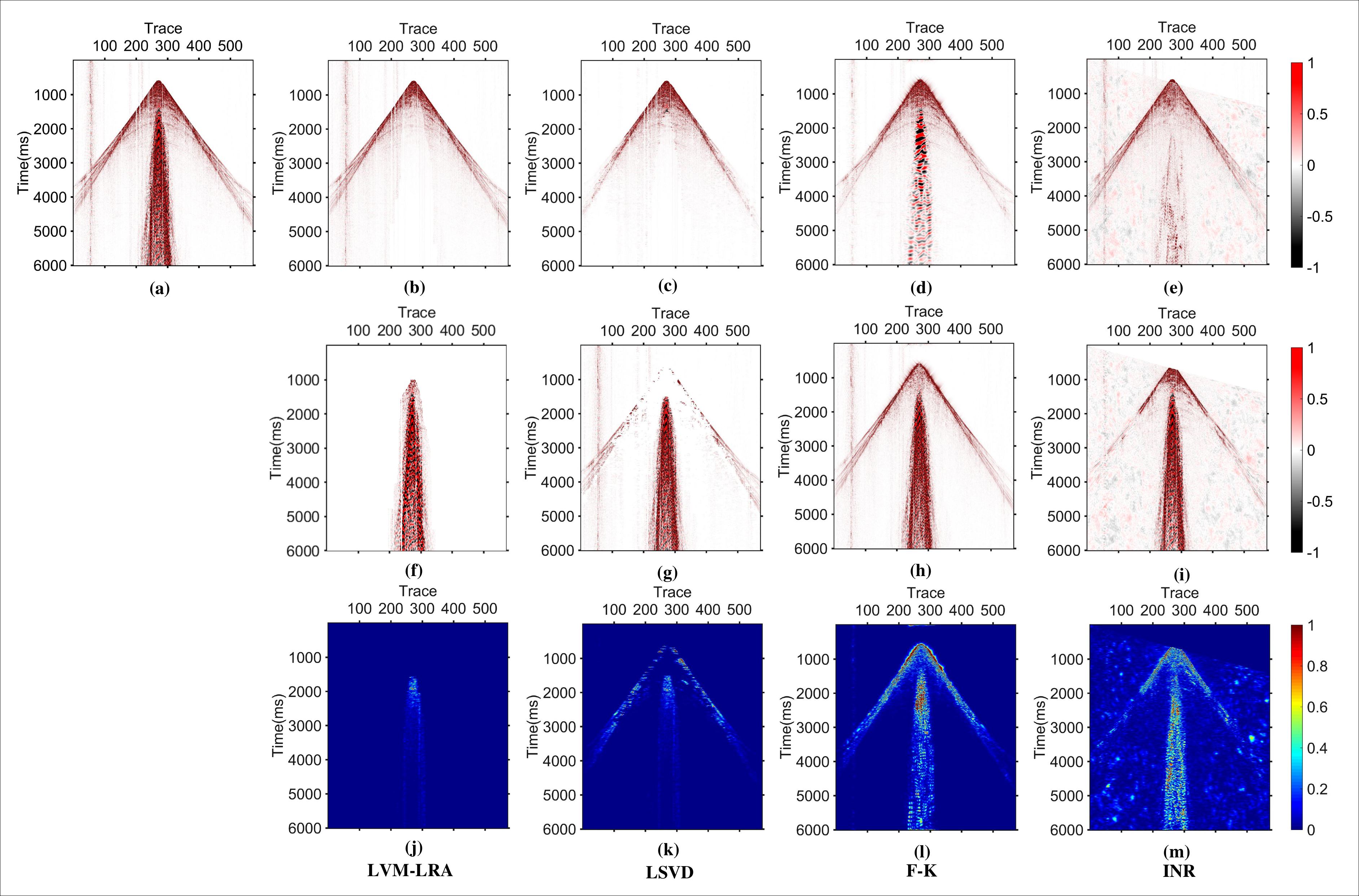}
	\caption{Denoising comparison on the second real seismic data. (a) Original seismic data. Denoised results, removal noise, and local similarity comparisons using (b)(f)(j) proposed LVM-LRA, (c)(g)(k) LSVD, (d)(h)(l) F-K, and (e)(i)(m) INR.}
	\label{Fig:syn5}
\end{figure*}

\subsection{Hyperparameter Configurations}
\label{subsec:hparams}

After establishing the LVM-LRA framework described in Section~\ref{sec:method}, two categories of parameters must be specified before implementation: the Stage~I parameters associated with prompt design in LVM-based mask generation, and the Stage~II parameters associated with the mask-guided low-rank decomposition, including the nuclear-norm weights and the ADMM solver settings.

In Stage~I, five text prompts and one visual prompt are used. The text prompts are further divided into three morphological prompts and two physical-property prompts, i.e., $N_m=3$ and $N_p=2$, whereas $N_v=1$ denotes a representative visual exemplar of ground roll. In this study, the three morphological prompts are specified as ``fan-shaped, low-frequency, coherent seismic events near the surface,'' ``cone-shaped, laterally continuous seismic energy with smooth wavefronts near the surface,'' and ``low-slope, coherent surface events forming broad fan-like patterns.'' The two physical-property prompts are specified as ``surface-wave noise with low apparent velocity and dispersion'' and ``near-surface guided-wave energy with strong amplitude and dispersive behavior,'' respectively.

As shown in \eqref{equ:stage_model}, after normalizing $\mathbf{Y}$, the nuclear-norm regularization weights are set to $\lambda_{\mathrm{g}}=1.0\times10^{-2}$ and $\lambda_{\mathrm{s}}=5.0\times10^{-3}$. The ADMM penalty parameter is set to $\rho_1=\rho_2=\rho_3=
3$, and the maximum number of iterations is set to $K_{m}=200$. The stopping tolerance is set to $\varepsilon=10^{-4}$. Unless otherwise specified, the same parameter settings are used throughout all experiments.

\section{Results}
\label{sec:results}
\subsection{Validation on Synthetic Seismic Data}
\label{subsec:syn}

In the synthetic experiment, a clean seismic reference is used, and a strong ground-roll component is superimposed to generate the contaminated input with an SNR of 1.45~dB, as shown in Fig.~\ref{Fig:syn0}(a). The proposed LVM-LRA method is compared with three representative baselines: LSVD, F--K, and INR. Figs.~\ref{Fig:mask}(a) and (b) show the ground-roll localization masks estimated by LVM-LRA and LSVD for the synthetic example. The mask produced by LVM-LRA is more compact and concentrated within the ground-roll-dominant cone, whereas the LSVD mask exhibits a less precise support region and includes more responses outside the main ground-roll area. This comparison provides an intuitive explanation for the subsequent separation results.

Fig.~\ref{Fig:syn0} summarizes the recovered useful-image sections, the extracted ground-roll components, and the local similarity maps between the reconstructed results and the extracted ground roll. From the recovered useful images in Figs.~\ref{Fig:syn0}(b)--(e), together with the corresponding extracted ground-roll sections in Figs.~\ref{Fig:syn0}(f)--(i), the proposed LVM-LRA method preserves the most coherent events while effectively suppressing the dominant ground-roll fan. By contrast, LSVD, F--K, and INR all exhibit varying degrees of residual interference and distortion in the overlap region between the useful image and ground roll.

Additional evidence is provided by the local similarity maps in the bottom row of Fig.~\ref{Fig:syn0}. The local similarity map produced by LVM-LRA exhibits high responses only in a limited portion of the ground-roll-dominant cone, indicating weaker residual ground-roll interference and more complete separation. In contrast, the baseline methods produce broader, more fragmented, and more diffuse similarity patterns in the central overlap region, suggesting less stable separation under strong signal--noise overlap.

\begin{figure*}[t!]
	
	\vspace{-0.4cm}
	\centering
	\includegraphics[width=0.75\textwidth,  trim=15 15 15 15,
  clip]{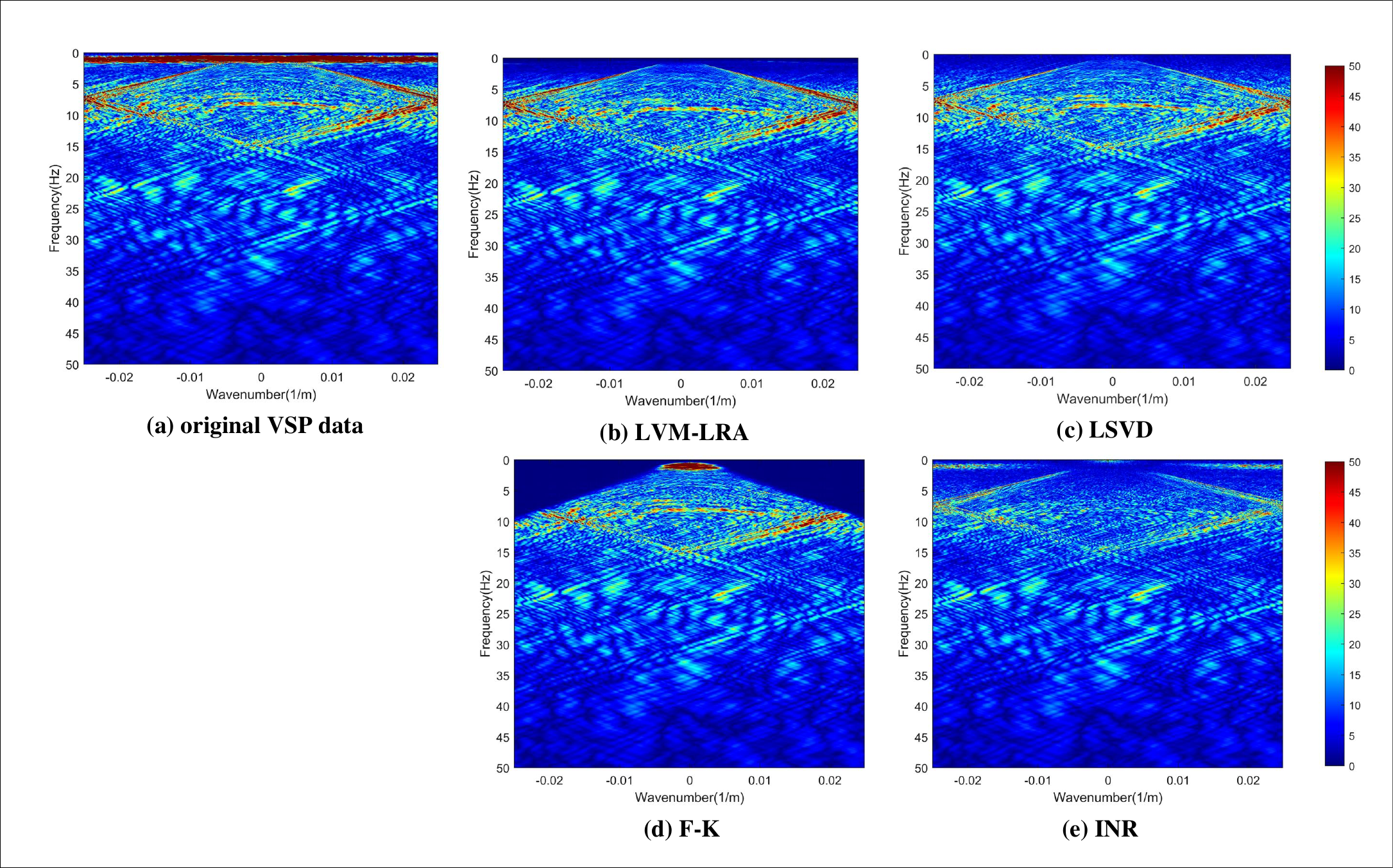}
	\caption{Spectral comparison of the denoised results in Fig. \ref{Fig:syn5}: (a), (b), (c), (d), and (e) Frequency–wavenumber spectra of Fig. \ref{Fig:syn5} (a), (b), (c), (d), and (e), respectively.}
	\label{Fig:syn6}
\end{figure*}

The superiority of LVM-LRA is further corroborated in the frequency--wavenumber domain, as shown in Fig.~\ref{Fig:syn1}. The spectrum of the original data in Fig.~\ref{Fig:syn1}(a) exhibits pronounced low-velocity ground-roll energy distributed away from the compact useful-signal band. After processing, the spectrum of the LVM-LRA result in Fig.~\ref{Fig:syn1}(b) becomes more compact, with noticeably reduced dispersed energy associated with ground roll. By comparison, the spectra of LSVD, F--K, and INR in Figs.~\ref{Fig:syn1}(c)--(e) retain more residual energy outside the main useful-signal band, implying less complete suppression of ground-roll contamination. This transform-domain observation is consistent with the section-based comparisons shown in Fig.~\ref{Fig:syn0}.

In addition, a trace-level comparison is presented in Fig.~\ref{Fig:syn2} to assess waveform fidelity within the useful-image-dominant window. The traces reconstructed by LVM-LRA closely follow the ground truth in both event timing and waveform shape. By contrast, the traces reconstructed by LSVD, F--K, and INR exhibit more visible waveform deviations, including spurious oscillations and local amplitude distortions on several traces, reflecting residual ground-roll interference and imperfect preservation of the useful image.

To complement the qualitative observations, quantitative results are reported in Table~\ref{table:SNR} using the metrics defined in Section~\ref{subsec:metrics}. Among all compared methods, LVM-LRA achieves the highest SNR and the lowest local similarity values, indicating the most favorable tradeoff between ground-roll attenuation and useful-image preservation. These quantitative results are consistent with the visual comparisons in Fig.~\ref{Fig:syn0}, the mask comparison in Fig.~\ref{Fig:mask}, the spectral evidence in Fig.~\ref{Fig:syn1}, and the trace-level analysis in Fig.~\ref{Fig:syn2}.



\begin{table}[t]
	\renewcommand{\arraystretch}{1.3}
	\centering
	\caption{Performance Comparison of Synthetic Seismic Data Based on SNR and Local Similarity Metrics}  
	\label{table:SNR} 
	\centering
	\begin{center}  
		\begin{tabular}{|l|c|c|c|}  
			\hline 
			\multirow{2}{*}{Methods} & \multirow{2}{*}{SNR(dB)} & \multicolumn{2}{c|}{Local Similarity} \\
			\cline{3-4}
			& & Average & Variance \\
			\hline\hline      
            {INR}  & $-03.34$  & $0.0834$ & $0.0352$\\ 			
			\hline   
			{F-K} & ${07.54}$  &  $0.0897$  & $0.0397$    \\ 
			\hline  
            {LSVD}  & $02.55$  & $0.0699$ & $0.0456$  \\ 
			\hline  
			\textbf{Proposed LVM-LRA} & $\textbf{14.50}$  & $\textbf{0.0285}$  & $\textbf{0.0100}$   \\ 
			\hline  
		\end{tabular}  
	\end{center} 
\end{table}

\subsection{Validation on the First Real Seismic Data}
\label{subsec:vsp}

To further evaluate the robustness of the proposed LVM-LRA framework, experiments are conducted on the first field dataset. Since no clean reference is available, the evaluation mainly relies on the reconstructed useful-image sections, the extracted ground-roll components, the corresponding local similarity maps, and the estimated localization masks.

\begin{figure}[t!]
	
	\vspace{-0.4cm}
	\centering
	\includegraphics[width=0.5\textwidth,  trim=5 5 5 5,
  clip]{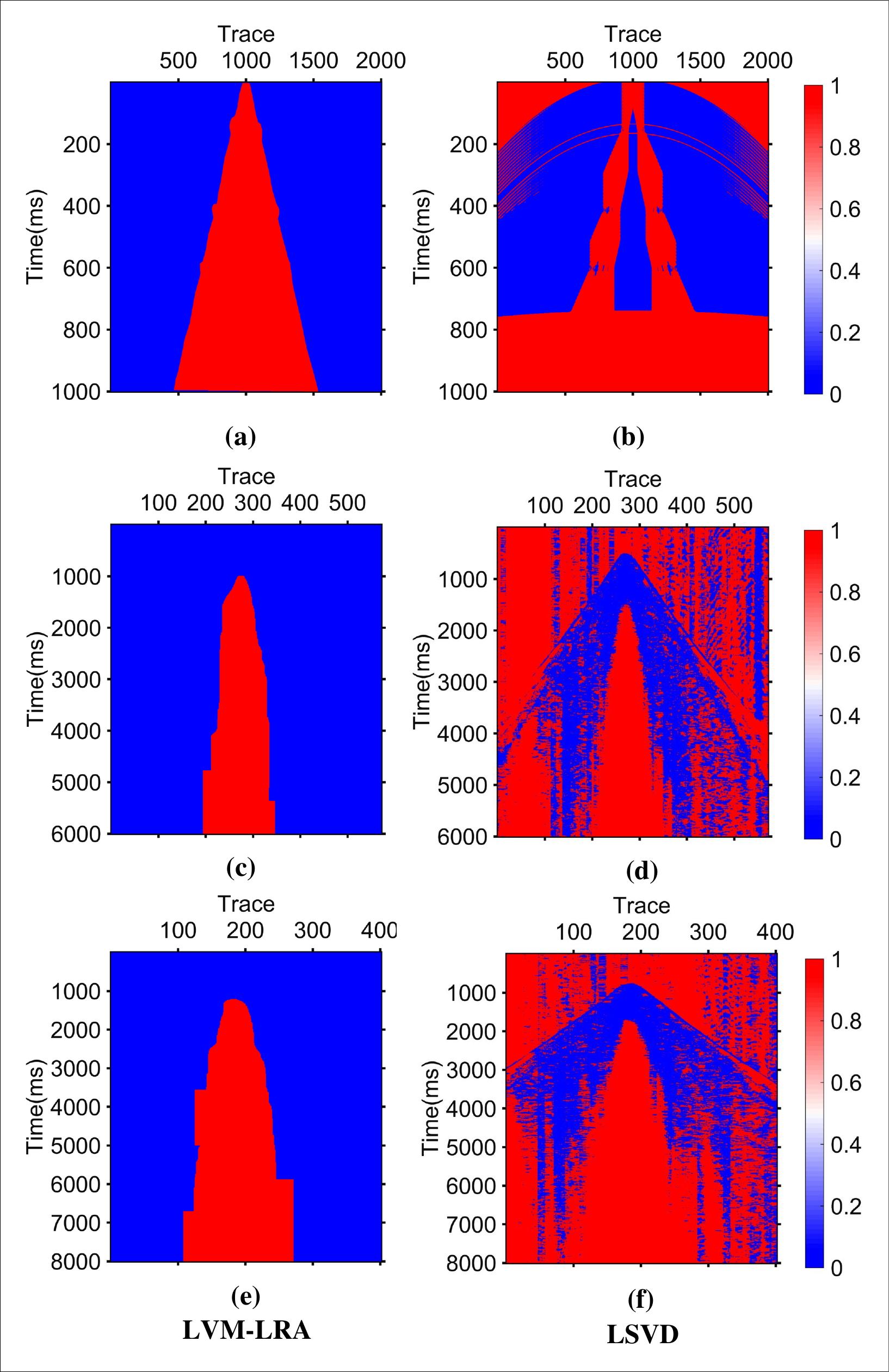}
	\caption{Ground-roll localization masks for the synthetic dataset, the first field dataset, and the second field dataset, given by LVM-LRA in (a), (c), and (e), respectively, and by LSVD in (b), (d), and (f), respectively.}
	\label{Fig:mask}
\end{figure}

Figs.~\ref{Fig:mask}(c) and (d) compare the ground-roll localization masks produced by LVM-LRA and LSVD. The mask generated by LVM-LRA is more compact and mainly concentrated in the ground-roll-dominant region, whereas the LSVD mask contains more spurious responses outside the main cone. This difference is consistent with the subsequent separation results.

Fig.~\ref{Fig:syn3} presents the processing results of LVM-LRA and three baselines: LSVD, F--K, and INR. Specifically, Figs.~\ref{Fig:syn3}(a)--(e) show the reconstructed useful-image sections, Figs.~\ref{Fig:syn3}(f)--(i) show the extracted ground-roll components, and Figs.~\ref{Fig:syn3}(j)--(m) present the local similarity maps between the reconstructed useful image and the extracted noise. Compared with the baseline methods, LVM-LRA yields clearer structural continuity, while the extracted ground roll is mainly concentrated within the dominant cone region. Moreover, its local similarity map exhibits weaker responses in the central overlap zone, indicating reduced signal leakage and more effective separation.

Further evidence is provided by the $f$--$k$ spectra in Fig.~\ref{Fig:syn4}. After processing, the spectrum of the LVM-LRA result becomes more concentrated, with reduced dispersed energy associated with low-velocity ground roll. By contrast, LSVD, F--K, and INR retain more residual energy outside the main useful-signal band. These observations are consistent with the section-based comparisons and the quantitative results in Table~\ref{table:AVE}.

\subsection{Validation on the Second Real Seismic Data}

To further verify the adaptability of the proposed LVM-LRA framework to different structural patterns in field data, experiments are conducted on a second real dataset, whose events exhibit a relatively narrower dip range and a more compact geometric distribution than those in the first field example. Since no clean reference is available, the evaluation is based on the reconstructed useful-image sections, the extracted ground-roll components, the local similarity maps, and the estimated localization masks.

Figs.~\ref{Fig:mask}(e) and (f) compare the ground-roll localization masks produced by LVM-LRA and LSVD. The mask generated by LVM-LRA remains compact and mainly confined to the ground-roll-dominant cone, whereas the LSVD mask contains more dispersed and spurious responses outside the main support region. This difference suggests that the proposed method can still provide a more reliable localization prior under a different structural configuration.

Fig.~\ref{Fig:syn5} presents the processing results of LVM-LRA and three baselines: LSVD, F--K, and INR. Specifically, Figs.~\ref{Fig:syn5}(a)--(e) show the reconstructed useful-image sections, Figs.~\ref{Fig:syn5}(f)--(i) show the extracted ground-roll components, and Figs.~\ref{Fig:syn5}(j)--(m) present the local similarity maps between the reconstructed useful image and the extracted noise. Compared with the baseline methods, LVM-LRA better preserves the main structural features while confining the removed energy more closely to the ground-roll-dominant region. Its local similarity map also exhibits weaker responses in the central cone, indicating reduced signal leakage and more complete separation.

Further evidence is provided by the $f$--$k$ spectra in Fig.~\ref{Fig:syn6}. After processing, the spectrum of the LVM-LRA result becomes more compact, with less dispersed energy associated with low-velocity ground roll. By contrast, LSVD, F--K, and INR retain more residual energy outside the main useful-signal band. These observations are consistent with the section-based comparisons in Fig.~\ref{Fig:syn5}. The quantitative results in Table~\ref{table:AVE} further support this conclusion. LVM-LRA achieves the lowest local similarity values among all compared methods, indicating reduced signal leakage and more effective separation. Taken together, these results demonstrate that the proposed method maintains stable separation performance even when the structural characteristics of the field data differ from those of the first real example.

\subsection{Limitation}
A major limitation of the proposed method lies in its strong reliance on the Stage-I LVM for identifying ground-roll-dominant regions and generating the corresponding segmentation masks. When ground roll spreads over a broader fan-shaped area, its morphological patterns, textural characteristics, and spatial distribution may become increasingly similar to those of the useful image, making the two components more difficult to distinguish \cite{zhou2025segment}. As a result, the LVM-generated mask may exhibit under-segmentation, over-segmentation, or boundary inaccuracies \cite{lin2025large}, thereby reducing its effectiveness as a support prior for the subsequent separation stage.

To address this issue, future work may extend the current one-shot framework into a feedback-driven iterative refinement scheme \cite{tang2021recurrent}. Specifically, the ground-roll estimate obtained from Stage II could be used to refine the Stage-I mask, and the updated mask could then guide a new round of separation. Such an alternating process between mask estimation and component separation may improve the robustness of the proposed method when the initial segmentation is inaccurate.

\begin{table}[t]
	\renewcommand{\arraystretch}{1.3}
	\centering
	\caption{Performance Comparison of Real Seismic Data Based on Local Similarity Metrics}  
	\label{table:AVE} 
	\centering
	\begin{center}  
		\begin{tabular}{|l|c|c|c|c|}  
			\hline 
			\multirow{2}{*}{Methods} & \multicolumn{2}{c|}{Real Seismic Data \uppercase\expandafter{\romannumeral 1}} & \multicolumn{2}{c|}{Real Seismic Data \uppercase\expandafter{\romannumeral 2}} \\
			\cline{2-5}
			& Average & Variance & Average & Variance \\
			\hline\hline     
            {INR}  & $0.0493$  & $0.0164$ & $0.2179$  & $0.0635$\\ 
			\hline  
			{F-K} & $\textcolor{black}{0.0481}$  &  $0.0111$  & $0.1571$ & $0.0415$   \\ 
			\hline     
            {LSVD}  & $0.0101$  & $0.0028$ & $0.0092$  & $0.0026$\\ 
			\hline  
			\textbf{Proposed LVM-LRA} & $\textbf{0.0057}$  & $\textbf{0.0019}$  & $\textbf{0.0049}$ & $\textbf{0.0015}$   \\ 
			\hline  
		\end{tabular}  
	\end{center} 
\end{table}

\section{Conclusion}
\label{sec:conclusion}

In this work, we propose an LVM-LRA framework for ground-roll suppression in seismic gathers, formulated as a training-free wavefield separation approach. A promptable mask is first obtained from an LVM by converting the seismic gather into a visual representation and identifying the ground-roll-dominant region through text prompts and a lightweight visual exemplar. The estimated mask is then incorporated into a mask-guided low-rank decomposition stage formulated with dual low-rank regularization. Specifically, a mask-constrained local low-rank regularization term in the frequency domain is imposed on the ground-roll component, whereas a global low-rank regularization term in the frequency domain is imposed on the useful-image component. The resulting optimization problem is solved efficiently using an ADMM-based procedure with fixed hyperparameter settings, enabling stable separation without requiring pixel-level annotations or paired clean/noisy training data. Extensive experiments on synthetic and real datasets demonstrate that the proposed LVM-LRA method achieves a favorable balance between ground-roll attenuation and useful-image preservation and consistently outperforms representative baseline methods in both image-domain evaluations and frequency--wavenumber spectra.



\ifCLASSOPTIONcaptionsoff
\newpage
\fi

\bibliographystyle{IEEEtran}
\bibliography{groundroll}

\clearpage

\end{document}